# A Diversity-Aware Domain Development Methodology

Mayukh Bagchi[1]

[1]*Department of Information Engineering and Computer Science (DISI), University of Trento, Via Sommarive, 9 I-38123 Povo (TN), Italy.*

**Abstract**
The development of domain ontological models, though being a mature research arena backed by well-established methodologies, still suffer from two *key shortcomings*. Firstly, the issues concerning the *semantic persistency* of ontology concepts and their *flexible reuse* in domain development employing existing approaches. Secondly, due to the *difficulty* in understanding and reusing top-level concepts in existing foundational ontologies, the *obfuscation* regarding the semantic nature of domain representations. The paper grounds the aforementioned shortcomings in *representation diversity* and proposes a *three-fold* solution - (i) a pipeline for rendering concepts *reuse-ready*, (ii) a first characterization of a *minimalistic* foundational knowledge model, named *foundational teleology*, semantically explicating foundational distinctions enforcing the static as well as dynamic nature of domain representations, and (iii) a *flexible, reuse-native* methodology for *diversity-aware* domain development exploiting solutions (i) and (ii). The preliminary work reported validates the potentiality of the solution components.

## 1. Introduction

The emphasis on shared in the definition of a formal ontology implies that it is (ideally) to be *reused* by the community of practice which shares, albeit *partially*, a *similar conceptualization*. It has, however, been corroborated in several established studies (e.g., [1]) that reusing ontologies *"does not seem to be widespread"* and even if reused, it is *"not a consolidated practice"* [1] and is decided on an *ad-hoc* basis without any shared best practice. The present research project is positioned in this context of domain ontological model development grounded in reuse, wherein *domain* is taken to be defined as *"an ontological base that reveals an underlying teleology"* [2].

Let us now concentrate, at a high level, on three specific research challenges afflicting the aforementioned research context. Firstly, that of the status quo of *concept reuse* from *general purpose ontologies*. Clearly, as shown in [1], existing ontology concept reuse approaches suffer from mainly, but not only, the *semantic heterogeneity* arising out of the *mismatch* between the intended and the available conceptualization(s), even within the same domain. Secondly, the reality that *foundational ontologies*, despite having *"a great potential for reuse"*, are practically considered *"hard to understand and, consequently, difficult to (re)use"* [3], resulting in *obfuscation* of top-level distinctions especially while modelling (static and) *dynamic* domains (i.e., domains modelling functions as well as *actions*). Thirdly, the fact that there is no single methodology which, while modelling both static and dynamic domains, is founded in ontology *concept reuse*.

The proposal puts forth a representation *diversity-aware* approach [4] towards tackling the above research challenges, and thus, grounds itself in the assertion that representation diversity should be considered as a, quoting [5], *"feature which must be maintained and exploited"* and not as a *"defect that must be absorbed"*. Accordingly, the paper proposes a novel *three-fold* solution proposal as an ordered response to the three research challenges. The first solution component is to design and implement a pipeline rendering concepts from reusable general purpose ontologies *reuse-ready*. The second solution component entails the design of a minimalistic foundational knowledge model enforcing meaning to concepts in both static and dynamic domain representations. Thirdly, incorporating the above two solution components in a reuse-native, *diversity-aware* methodology for developing domain ontological representations.

The rest of the paper is organized as follows. Section 2 and 3 elaborates on the state-of-the-art and research objectives. Section 4 discuses the solution proposal and validation strategy. Section 5 concludes the paper by enunciating the preliminary work done.

## 2. State Of The Art

As from the introduction, the three research challenges directly map to state-of-the-art in: *ontology concept reuse*, *foundational ontologies* and *ontology development methodologies*.

The three types of ontology reuse - direct, indirect and hybrid - were exemplified in works such as, e.g., [1]. The study in [6] examines in detail the feasibility of ontology reuse and posits requirements for designing ontology reuse methodologies. The work on content ontology design patterns [7] is also founded in reuse. Grounded in the above work on ontology reuse is the specific aspect of *ontology concept reuse*, majorly observed with respect to highly domain-specific ontologies such as in biomedicine, e.g., [8] and agriculture, e.g., [9]. The more recent empirical work on BioPortal ontologies, e.g., [10] conclusively observes that both / *reuse and reusability is significantly low*, barring few exceptions. The observations are similar for ontologies in AgroPortal with the recent study [9] observing *"most ontologies overlap with, reuse, or map less than 5% (...) to other ontologies"*. Even concepts from general purpose ontologies in repositories like LOV [11] are hardly reused with the exception of, however, the W3C endorsed ontologies.

Next, let us focus on the landscape of foundational ontologies (*aka* upper or top-level ontologies). The review in [12] analyze seven upper ontologies according to a select set of software engineering criteria, amongst which we brief below few of the most used ones. DOLCE [13] is a comprehensive upper *"ontology of particulars"* grounded in the fundamental distinction between *endurants* and *perdurants*. It, however doesn't model *functions* or *roles* in its core taxonomy [14]. BFO [15] is a bi-ontological theory with two components - a *Snap* ontology of endurants and a *Span* ontology of perdurants, and is, almost solely, employed within the biomedical community. SUMO [16] acts a foundation for domain ontologies and contains about thousand terms grounded in the distinction between physical and abstract entities. UFO [17] is composed of three distinct ontology fragments - endurants, perdurants and social entities.

Finally, let us concentrate briefly on ontology development methodologies. The survey in

[18] provides an analysis of several early generation methodologies. METHONTOLOGY [19] proposed a *"life cycle to build ontologies based in evolving prototypes"*. Ontology Development 101 [20], instead, offered the flexibility of choosing top-down, bottom-up or middle-out approaches in engineering ontologies. More recently, the NeOn methodology [21] offers a set of *very generic* scenarios for reuse, re-engineering and merging of ontological resources. The eXtreme Design (XD) methodology [22], on the other hand, is very specific, in the sense that it is grounded on reusage of content ontology design patterns for modelling new ontologies.

## 3. Problem Statement

The current work is based on the *stratified nature* of *representation diversity* [4][1]. The starting point is the issue of semantic heterogeneity in domain representations which, as from the introduction, should be accommodated as a feature. To facilitate such an accommodation, the novel approach in [4] restates semantic heterogeneity as a problem of representation diversity stratified into four *characteristically autonomous* yet *functionally linked* representation layers:

- *Concept [Diversity] (L1)*, arising out of the many-to-many mapping between real world objects and their perceived representation as concepts [4, 23]
- *Language [Diversity] (L2)*[2], arising out of the many-to-many mapping between concepts and the words employed for their linguistic rendering due to linguistic phenomena such as polysemy and synonymy [24]
- *Knowledge [Diversity] (L4)*, arising out of the many-to-many mapping between entity types (etypes from now on) and the properties employed to model them [4, 25], and
- *Data [Diversity] (L5)*[3]., arising out of the many-to-many mapping between entities and the property values employed to describe them [4].

Grounded in the above foundation, the research question stands: "How can *concepts* from reusable general purpose ontologies in different *languages* be *methodologically* reused to model a *reusable and shareable* domain (knowledge) representation for integrating data exhibiting different genres of *semantic heterogeneity*?" It can be instantiated into the following *independent* but *related* research objectives:

**(O1)** *To design and implement a pipeline rendering concepts from reusable 'general purpose' ontologies reuse-ready* (accommodating L1, L2 (aka L1,2) diversity)
**(O2)** *To design a foundational knowledge model explicating foundational distinctions which accommodate static and dynamic domain representations*
**(O3)** *To design and implement a diversity-aware methodology for modelling domain ontological representations exploiting (O1) and (O2)* (accommodating L4 diversity)

---

[1]See [4]. Details omitted due to lack of space.
[2]L3 out of scope for this work.
[3]Detailed deliberation on L5 out of scope for the current proposal.

## 4. Solution Approach

The solution proposal introduces the three solution components, namely subsections: **(4.1)**, **(4.2)** and **(4.3)**, corresponding to the three research objectives - **(O1)**, **(O2)** and **(O3)** respectively. Further, the validation strategy of the overall solution framework is also discussed.

### 4.1. Making Concepts Reuse-Ready (O1)

At the outset, let us explain the following terms which are crucial for understanding the *curated* pipeline for making ontology concepts reuse-ready:

- **Synset** - *Synsets* [26] are sets of synonyms used to represent word senses
- **Word Sense Rank (WSR)** - The rank of a synonym in a synset, with the preferred term having WSR 1.

**UKC** - UKC stands for Universal Knowledge Core [27, 24]. It is a multilingual lexical-semantic resource composed of two components - Concept Core (CC) and Language Core (LC). The CC is a semantic network where the nodes are *alinguistic concepts* structured employing semantic relations (such as hypernym-hyponym), with each concept uniquely identified *only* by a Global IDentifier (GID). In essence, the semantic network is a Directed Acyclic Graph (DAG) codifying the space of possibilities of all existing and forthcoming concepts, and can be considered as *background knowledge*. The LC, on the other hand, is a union of language-specific modules, with each module being the set of *"words, senses, synsets, glosses and examples"* [27] in a specific language existent or upcoming in the UKC. Semantically equivalent synsets across languages are univocally interconnected and represented by a single GID.

Our proposed pipeline for rendering ontology concepts reuse-ready is sequentially enumerated as follows (corresponding to each step in figure (1) -

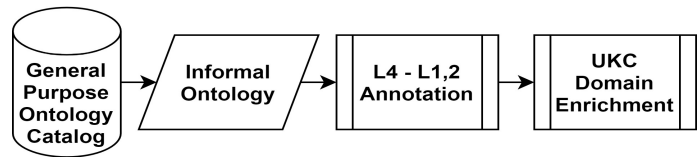

Figure 1: Pipeline for Making Concepts Reuse-Ready

1. *General Purpose Ontology Catalogue* - We restrict ourselves to high quality reusable general purpose ontologies from LOV [11].
2. *Informal Ontology Selection* - This step is to select an ontology from the the LOV. The strategy is to select reusable ontologies, for a particular domain, based on the *number of incoming links* for a certain ontology in LOV, which, being a popularity metric, provide a measure of how many other ontologies have reused it (either partially or fully). The selected ontology is termed as *informal* as the concepts it expresses are yet to be rendered *alinguistic*.

3. *L4 - L1,2 Annotation* - The central step of the pipeline performed by a *domain expert*, wherein each concept from the selected informal ontology is either annotated with its GID if it is already present in the UKC CC, or a new concept with a new GID is created in the CC if the concept is new with respect to the CC. Concretely, the sub-steps are as follows -
    a) Each concept from the informal ontology's class hierarchy, object property hierarchy and data property hierarchy is considered (sequentially; one hierarchy at a time) in a top-down order, and its sense is understood from the gloss provided in the annotation properties (mostly captured in *rdfs:comment* and/or *rdfs:isDefinedBy*).
    b) The concept is semantically searched in the UKC CC via the LC interface matching with the natural language in which the concept label is expressed. The search results in one of the following two scenarios:-
        i. *(S1): Synonymous Match* between the ontology concept and an existing concept found in the UKC CC.
        ii. *(S2): No Synonymous Match* between the ontology concept and any potential concept in the UKC CC.
    c) Concurrently with step 3(a) and 3(b), a UKC-compatible spreadsheet capturing requisite information is generated. The principle information to be recorded in the spreadsheet are as follows:-
        i. *(S1)*: In case of a *Synonymous Match*, the GID and the Word Sense Rank of the concept have to be recorded in the spreadsheet, alongside its parent concept (and its GID).
        ii. *(S2)*: In case of *No Synonymous Match*, a negative integer is recorded in place of the concept's GID. For all such concepts, the recorded negative integer should follow a decremental (negative integer) sequence starting from '-1'. In addition, the parent concept (and its GID) is also recorded. Further, for this scenario in specific, the gloss of the concept should also be recorded in the spreadsheet (respecting *Genus-Differentia* paradigm [28]).
4. *UKC Domain Enrichment* - The spreadsheet is imported into the UKC CC via the spreadsheet importer API, wherein post import, two highlights are crucial - (i) all the new concepts (annotated with negative integers) now have their own, unique GIDs, and (ii) one more concept hierarchy gets formalized within the DAG of the CC, that of the imported ontology. Thus, the concepts are finally *reuse-ready*.

### 4.2. Foundational Teleology (FT) (O2)

The solution that the proposal puts forth for **(O2)** is a *foundational teleology* where teleology is an ontology that, quoting [29], *"focus on function and on how a chosen representation fits a certain purpose, this being the basis for a general model for the diversity of knowledge"*. An initial characterization of the foundational teleology (FT) can be found in figure (2) above, where the ovals stand for top-level concepts, the lines for subsumption relation and the dashed lines for relations. The teleology is grounded in the theory of teleosemantics [29] and models ends and goals (via functions and actions). Further, we speak of a foundational teleology in the sense of a teleology explicating the *foundational distinctions* and *foundational relations* which are central for semantically grounding the static and dynamic nature of a 'domain' ontological model.

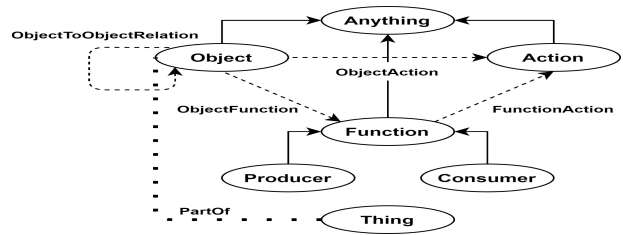

**Figure 2: Foundational Teleology (FT)**

Let us now focus on the composition of the foundational teleology in terms of its foundational distinctions and foundational relations. The root of the FT is *Anything* which we define as the *substratum* [30] for modelling purpose-driven domain representations. It is a bearer of properties but has its identity independent from them, thereby exhibiting permanence and transcending space-time. *Anything* is specialized into *Object, Function* and *Action* - the core foundational distinctions which model the function-focused and purpose-driven nature of domains, according to the theory of teleosemantics [29]. Objects are defined as *representations of what is perceived* (such as a person). Functions model *the expected behavior of objects* in terms of it performing a certain set of actions, wherein Actions *represent how objects change in time* (i.e., functions and actions model the dynamics of a domain). For example, a person can have several functions such as father, professor etc., each of which can be modelled as a set of (admissible) actions (such as teach, evaluate etc. for professors). Further, functions, from teleosemantics, are specialized into - *Producer* and *Consumer*, wherein producer is an object performing an action affecting another object, where the 'another' object is the consumer. *Thing*, instead, is the domain reference context. There are four foundational relations interconnecting the above distinctions. The *ObjectToObjectRelation* captures the intra-relationship between objects. The relation between object and function, function and action, and object and action are modelled by *ObjectFunction, FunctionAction*, and *ObjectAction* repsectively.

### 4.3. The Diversity-Aware Domain Development Methodology (O3)

Finally, the domain development methodology (figure 3) is elucidated. It is made possible due to the integration of the two previous solution components, namely **(4.1)** and **(4.2)**, in the framework of a single general methodology. Let us examine each step of the methodology corresponding to the steps in figure 3.

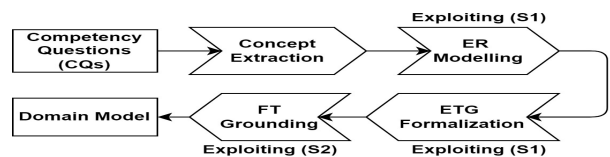

**Figure 3: The Diversity-Aware Domain Development Methodology Methodology**

**(1) Competency Questions (CQs) -** CQs [31] are *input* with reference to the methodology. They are generated by a comprehensive process (detailed in [32]) elaborating the users of the domain model the potential questions they want to ask from the domain model in certain scenarios.

**(2) Concept Extraction -** Concept extraction refers to the elicitation of concepts from CQs (from step (1)) and as such is a semantics-intensive task. The study in [33] confirms that concept elicitation from CQs, till date, is a heuristics-driven task and has no dedicated mechanism. We propose below a (domain expert performed) mechanism, grounded in faceted knowledge organization [34, 35], for eliciting concepts from CQs -

1. **Raw CQ -** CQ in natural language as obtained from step (1).
2. **Kernel CQ -** Raw CQ minus all the auxillary/apparatus words (such as stop words) resulting in each term denoting a concept. Further, this step also elicits the *latent concepts* [34], concepts which are explicitly hidden but are *implied or suggestive*.
3. **Analyzed CQ -** Each concept in Kernel CQ classified as common, core or contextual, where: (i) *common* refers to concepts from *space* and *time* [34]; (ii) *core* refers to concepts fundamental to the domain to be modelled (for e.g., tourist facilities), and (iii) *contextual* refers to concepts which are *highly localized* with respect to the *reference context* (for e.g., *malga* as a tourist facility in Italy)
4. **Classified CQ -** Each concept in Analyzed CQ further classified as *object, action* or *function*
5. **Attributed CQ -** Each concept in Classified CQ enriched with requisite *object properties* and *data properties*.

**(3) ER Modelling -** Once concepts are extracted from CQs, we concentrate on modelling the entity-relationship (ER) model following the two phases below -

***Making Concepts Reuse-Ready*** - Here we exploit the pipeline for rendering concepts reuse-ready as proposed in **(4.1)**. From the LOV, we perform informal ontology selection resulting in ontologies whose concepts can be partially reused to model the ER (for e.g., ontologies on tourism and facilities for the domain 'tourist facilities'). Next, we perform L4 - L1,2 Annotation with respect to the selected informal ontologies and enrich the UKC, thus rendering the concepts formal (with unique GIDs) and reuse-ready.

***ER Development*** - Firstly, we specify the reference context - *Thing* - for which we want to develop the ER model (e.g., tourist facilities in Trentino, Italy from 01.01.2020-01.01.2021). We then instantiate the object hierarchy, the function hierarchy and the action hierarchy, respectively, with respect to *Thing*. The next step is to Interrelate *objects*, *functions* and *actions*, and subsequently with requisite object and data properties.

**(4) ETG Formalization -** In this step, we render the ER model as completely formal, by which we mean all concepts structuring it can be uniquely expressed using a GID. For achieving this, we implement the pipeline for rendering concepts reuse-ready for the second time, but, with two *key* differences - (i) this time, we consider the ER model we developed and hence don't perform an informal ontology selection from any catalog, and (ii) the objective being different, of rendering informal concepts as formal with GIDs. We call this formal ontological model as Entity Type Graph (ETG) [4].

**(5) FT Grounding** - Once the ER model has been fully formalized into an ETG, the final step is to ground the ETG to the foundational teleology (exploiting solution **(4.2)**). We propose to do this by aligning the object, function and action hierarchy to their respective foundational distinctions in the FT, and as a result, also grounding the relationships in the ETG into their semantically appropriate foundational relations.

**(6) Domain Model** - Finally, we have the domain ontological model, which is the principal output of the methodology. The proposal is to host the domain models developed in *Liveschema*[4] [36], which is our own meta-catalog of 'general purpose' ontologies.

**Validation** - The validation strategy for the solution components are briefly as follows: (i) For component **(4.1)**, quantitative evaluation following metrics adapted from [1]; (ii) For component **(4.2)**, qualitative evaluation following tried and tested principles of concept hierarchies as enunciated in [34], and (iii) For component **(4.3)**, testing, implementation and potential tuning of the methodology in KDI/KGE master degree course projects[5][6] and EU projects.

## 5. Preliminary Results

### 5.1. Work Done on (4.1)

In collaboration with international domain annotation experts, *L4 - L1,2 Annotation* was carried out on thirty most popular 'general purpose' ontologies, selected from the LOV catalog across categories like Geography, Time, Academia etc. Additionally, concepts from ten such annotated ontologies files were translated into the Hindi language.

### 5.2. Work Done on (4.2)

The emphasis of the preliminary work done regarding the solution component **(4.2)** was on designing a first characterization of the foundational teleology starting from the foundations of the theory of teleosemantics [29].

### 5.3. Work Done on (4.3)

Firstly, the methodology was designed and the series of papers in [37, 38, 39, 40] developed the stratified knowledge representation theory and the *cognitive grounding* behind the proposed methodology. The second aspect, which is a consequence of the proposed stratified knowledge representation theory, is the definition of a very compelling set of qualitative norms (see [37]) which allow for the construction of high quality domain models.

## Acknowledgements

I am grateful to my advisor Prof. Fausto Giunchiglia for supervising my research and for countless inspirational discussions. This PhD is funded by *DELPhi* project - MIUR (PRIN) 2017.

---

[4]http://liveschema.eu/
[5]http://knowdive.disi.unitn.it/teaching/kdi/
[6]https://unitn-knowledge-graph-engineering.github.io/KGE2022-website/

# References

[1] M. Fernández-López, M. Poveda-Villalón, M. C. Suárez-Figueroa, A. Gómez-Pérez, Why are ontologies not reused across the same domain?, Journal of Web Semantics 57 (2019).

[2] R. Smiraglia, Domain analysis for knowledge organization: tools for ontology extraction, Chandos Publishing, 2015.

[3] M. Fernández-López, A. Gómez-Pérez, M. C. Suárez-Figueroa, Methodological guidelines for reusing general ontologies, Data & Knowledge Engineering 86 (2013) 242–275.

[4] F. Giunchiglia, A. Zamboni, M. Bagchi, S. Bocca, Stratified data integration, in: 2nd International Workshop On Knowledge Graph Construction (KGCW), Co-located with the Extended Semantic Web Conference (ESWC) 2021, Online, 2021.

[5] F. Giunchiglia, Managing diversity in knowledge, in: IEA/AIE, 2006, p. 1.

[6] E. Simperl, Reusing ontologies on the semantic web: A feasibility study, Data & Knowledge Engineering 68 (2009) 905–925.

[7] A. Gangemi, V. Presutti, Ontology design patterns, in: Handbook on ontologies, Springer, 2009, pp. 221–243.

[8] M. R. Kamdar, T. Tudorache, M. A. Musen, A systematic analysis of term reuse and term overlap across biomedical ontologies, Semantic web 8 (2017) 853–871.

[9] A. Laadhar, E. Abrahão, C. Jonquet, Analysis of term reuse, term overlap and extracted mappings across agroportal semantic resources, in: International Conference on Knowledge Engineering and Knowledge Management, Springer, 2020, pp. 71–87.

[10] M. R. Kamdar, T. Tudorache, M. A. Musen, Investigating term reuse and overlap in biomedical ontologies, in: CEUR workshop proceedings, volume 1515, NIH Public Access, 2015.

[11] P.-Y. Vandenbussche, G. A. e. Atemezing, Linked open vocabularies (lov): a gateway to reusable semantic vocabularies on the web, Semantic Web 8 (2017) 437–452.

[12] V. Mascardi, V. Cordì, P. Rosso, A comparison of upper ontologies., in: Woa, volume 2007, Citeseer, 2007, pp. 55–64.

[13] C. Masolo, S. Borgo, A. Gangemi, N. Guarino, A. Oltramari, Wonderweb deliverable d17, Science Direct Working Paper No S1574-034X (04) (2002) 70214–8.

[14] J. Röhl, L. Jansen, Why functions are not special dispositions: an improved classification of realizables for top-level ontologies, Journal of Biomedical Semantics 5 (2014) 1–16.

[15] B. Smith, P. Grenon, L. Goldberg, Biodynamic ontology: Applying bfo in the biomedical domain, Studies in Health and Technology Informatics 102 (2004) 20–38.

[16] I. Niles, A. Pease, Towards a standard upper ontology, in: Proceedings of the international conference on Formal Ontology in Information Systems-Volume 2001, 2001, pp. 2–9.

[17] G. Guizzardi, R. de Almeida Falbo, R. S. Guizzardi, Grounding software domain ontologies in the unified foundational ontology (ufo): The case of the ode software process ontology., in: CIbSE, Citeseer, 2008, pp. 127–140.

[18] M. Fernández-López, Overview of methodologies for building ontologies, in: IJCAI99 Ontology Workshop, volume 430, Citeseer, 1999.

[19] M. Fernández-López, A. Gómez-Pérez, N. Juristo, Methontology: from ontological art towards ontological engineering (1997).

[20] N. F. Noy, D. L. McGuinness, et al., Ontology development 101: A guide to creating your

first ontology, 2001.
- [21] M. C. Suárez-Figueroa, A. Gómez-Pérez, M. Fernández-López, The neon methodology for ontology engineering, in: Ontology engineering in a networked world, Springer, 2012, pp. 9–34.
- [22] E. Blomqvist, K. Hammar, V. Presutti, Engineering ontologies with patterns-the extreme design methodology., Ontology Engineering with Ontology Design Patterns (2016) 23–50.
- [23] K. Janowicz, The role of space and time for knowledge organization on the semantic web, Semantic Web 1 (2010) 25–32.
- [24] F. Giunchiglia, K. Batsuren, G. Bella, Understanding and exploiting language diversity, in: *IJCAI*, 2017, pp. 4009–4017.
- [25] F. Giunchiglia, M. Fumagalli, Entity Type Recognition – Dealing with the Diversity of Knowledge, in: 17th KR Conference, 2020, pp. 414–423.
- [26] G. A. Miller, Wordnet: a lexical database for english, Communications of the ACM 38 (1995) 39–41.
- [27] F. Giunchiglia, K. Batsuren, A. Freihat, One world - seven thousand languages, in: 19th International Conference on Computational Linguistics and Intelligent Text Processing, Hanoi, Vietnam, 2018.
- [28] W. T. Parry, E. A. Hacker, Aristotelian logic, Suny Press, 1991.
- [29] F. Giunchiglia, M. Fumagalli, Teleologies: Objects, actions and functions, in: International conference on conceptual modeling, Springer, 2017, pp. 520–534.
- [30] J. Bennett, Substratum, History of Philosophy quarterly 4 (1987) 197–215.
- [31] M. Grüninger, M. S. Fox, The role of competency questions in enterprise engineering, in: Benchmarking—Theory and practice, Springer, 1995, pp. 22–31.
- [32] F. Giunchiglia, S. Bocca, M. Fumagalli, M. Bagchi, A. Zamboni, itelos-building reusable knowledge graphs, arXiv e-prints (2021) arXiv–2105.
- [33] D. Wiśniewski, J. Potoniec, A. Ławrynowicz, C. M. Keet, Analysis of ontology competency questions and their formalizations in sparql-owl, J. of Web Sem. 59 (2019) 100534.
- [34] S. R. Ranganathan, Prolegomena to Library Classification, Asia Publishing House (Bombay and New York), 1967.
- [35] S. R. Ranganathan, Colon classification, [6th ed.] ed., Asia Pub. House Bombay, New York, 1964.
- [36] M. Fumagalli, M. Boffo, D. Shi, M. Bagchi, F. Giunchiglia, Liveschema: A gateway towards learning on knowledge graph schemas, arXiv preprint arXiv:2207.06112 (2022).
- [37] F. Giunchiglia, M. Bagchi, Millikan + ranganathan – from perception to classification, in: 5th Cognition And Ontologies (CAOS) Workshop, Co-located with the 12th International Conference on Formal Ontology in Information Systems (FOIS) 2021, Bolzano, Italy, 2021.
- [38] F. Giunchiglia, M. Bagchi, Object recognition as classification via visual properties, in: 17th International ISKO Conference and Advances in Knowledge Organization, Aalborg, Denmark, 2022.
- [39] F. Giunchiglia, M. Bagchi, X. Diao, Visual ground truth construction as faceted classification, arXiv preprint arXiv:2202.08512 (2022).
- [40] F. Giunchiglia, M. Bagchi, Representation heterogeneity, in: 1st International Workshop on Formal Models of Knowledge Diversity (FMKD), Joint Ontology WOrkshops (JOWO), Jönköping University, Jönköping, Sweden, 2022.